\documentclass{article}
\usepackage{spconf,amsmath,graphicx}
\usepackage[fontsize=9pt]{scrextend}

\usepackage{booktabs}
\usepackage{balance}

\usepackage{amsmath,epsfig,amssymb,amsthm,url,amsfonts}
\usepackage{algorithm}
\usepackage{algpseudocode}
\usepackage{multirow}
\usepackage{mdframed}
\usepackage{url}
\usepackage{enumitem}
\usepackage[makeroom]{cancel}
\usepackage{dblfloatfix}
\usepackage{array}
\usepackage{comment}
\usepackage{epstopdf}
\usepackage{float}
\usepackage{subfig}
\usepackage{breqn}
\usepackage{cite}
\usepackage{xcolor}
\usepackage{cases}
\usepackage{tabularx}
\usepackage[printonlyused,withpage]{acronym}
\usepackage{balance}
\usepackage{graphbox} 

\usepackage{hyperref}
\setlist[itemize]{label=$\triangleright$}
\newtheoremstyle{break}
{}
{}
{\itshape}
{}
{\bfseries}
{.}
{\newline}
{}
\theoremstyle{break}

\theoremstyle{definition}

\newcommand{\vect}[1]{\mathbf{#1}}
\newcommand{\bs}[1]{\boldsymbol{#1}}
\newcommand{\E}{\mathbb{E}}
\usepackage{url}

\makeatletter
\def\thmhead@plain#1#2#3{%
	\thmname{#1}\thmnumber{\@ifnotempty{#1}{ }\@upn{#2}}%
	\thmnote{ {\the\thm@notefont#3}}}
\let\thmhead\thmhead@plain
\makeatother

\graphicspath{{./figs/}}

\newcommand{\argmax}{\operatornamewithlimits{argmax}}

\newcommand{\lk}{ \left\{ }
\newcommand{\rk}{ \right\} }

\newcommand{\Hb}{{\bf H}}

\newcommand{\bb}{{\bf b}}

\newcommand{\Wb}{{\bf W}}

\newcommand{\diag}{\mbox{{diag}}}

\newsavebox\mybox

\acrodef{SE}{speech enhancement}
\acrodef{STFT}{short-time Fourier transform}
\acrodef{STOI}{short-time objective intelligibility}
\acrodef{PSD}{power spectral density}
\acrodef{NMF}{non-negative matrix factorization}
\acrodef{AV}{audio-visual}
\acrodef{DNN}{deep neural network}
\acrodefplural{DNNs}{deep neural networks}
\acrodef{VAE}{variational auto-encoder}
\acrodefplural{VAEs}{variational auto-encoders}
\acrodef{CVAE}{conditional variational auto-encoder}
\acrodefplural{CVAEs}{conditional variational auto-encoders}
\acrodef{A-VAE}{audio VAE}
\acrodef{V-VAE}{visual VAE}
\acrodef{AV-CVAE}{audio-visual CVAE}
\acrodef{ROI}{region of interest}
\acrodef{MCMC}{Markov Chain Monte Carlo}
\acrodef{EM}{expectation-maximization}
\acrodef{MCEM}{Monte Carlo expectation-maximization}
\acrodef{TF}{time frequency}
\acrodef{ELBO}{evidence lower bound}
\acrodef{ROI}{region of interest}
\acrodef{LR}{Living Room}
\acrodef{SDR}{signal-to-distortion ratio}
\acrodef{PESQ}{perceptual evaluation of speech quality}
\acrodef{ASE}{audio speech enhancement}
\acrodef{VSE}{visual speech enhancement}
\acrodef{AVSE}{Audio-visual speech enhancement}
\acrodef{SNR}{signal-to-noise ratio}
\acrodefplural{SNRs}{signal-to-noise ratios}
\acrodef{LSTM}{long short-term memory}
\acrodef{DNNs}{deep neural networks}

\title{Audio-visual Speech Enhancement with a Deep Kalman Filter Generative Model}
\name{%
Ali Golmakani, %
Mostafa Sadeghi, and
Romain Serizel %
\thanks{Experiments presented in this paper were carried out using the Grid'5000 testbed, supported by a scientific interest group hosted by Inria and including CNRS, RENATER, and several Universities as well as other organizations (see https://www.grid5000.fr). }}
\address{%
Université de Lorraine, CNRS, Inria, LORIA, F-54000 Nancy, France}
\begin{document}
%
\maketitle
\begin{abstract}
Deep latent variable generative models based on variational autoencoder (VAE) have shown promising performance for audio-visual speech enhancement (AVSE). The underlying idea is to learn a VAE-based audio-visual prior distribution for clean speech data, and then combine it with a statistical noise model to recover a speech signal from a noisy audio recording and video (lip images) of the target speaker. Existing generative models developed for AVSE do not take into account the sequential nature of speech data, which prevents them from fully incorporating the power of visual data. In this paper, we present an audio-visual deep Kalman filter (AV-DKF) generative model which assumes a first-order Markov chain model for the latent variables and effectively fuses audio-visual data. Moreover, we develop an efficient inference methodology to estimate speech signals at test time. We conduct a set of experiments to compare different variants of generative models for speech enhancement. The results demonstrate the superiority of the AV-DKF model compared with both its audio-only version and the non-sequential audio-only and audio-visual VAE-based models.

\end{abstract}
\begin{keywords}
Audio-visual speech enhancement, generative model, variational autoencoder, deep Kalman filter.
\end{keywords}
\section{Introduction}
\label{sec:intro}
\ac{AVSE} is the task of estimating a clean speech signal given a noisy audio recording, as well as visual information (e.g., lip images) of the speaker \cite{michelsanti2021overview}. Visual data provide complementary information that could be very helpful for speech enhancement, especially when the audio recording is highly noisy \cite{michelsanti2021overview,kang2022expression}. Furthermore, visual data are robust with respect to acoustic noise and could help discriminate between the target speaker and potential concurrent speakers. Over the last decade and with the unprecedented progress made in deep learning, the AVSE problem has been extensively revisited \cite{ephrat2018looking,afouras2018conversation,gabbay2018visual,michelsanti2021overview}.

A dominant AVSE approach is to design and train a deep neural architecture that fuses audio and visual features, extracted from video and noisy audio data, respectively, to estimate the clean speech signal directly. This approach is data-driven and, as such, its success and generalization performance depend heavily on the amount of training data and their diversity, e.g., in terms of noise types. In contrast to this \textit{supervised} AVSE framework, a recent alternative  paradigm is to combine the classical model-based methods, e.g., maximum a posteriori (MAP) estimation, with the expressive power of \ac{DNNs} to perform \textit{unsupervised} AVSE \cite{sadeghi2020audio,sadeghi2021mixture,sadeghi2020robust,sadeghi2021switching}. More precisely, in a pre-training phase, the statistical characteristics of speech signals in the time-frequency domain are learned via a deep generative model based on variational autoencoders (VAEs) \cite{KingW14}, with only \textit{clean}
\ac{AV} data. The learned speech generative model, serving as a deep speech prior, is then combined with a parametric statistical model for noise, whose parameters along with the clean speech signal are estimated following an \ac{EM}-based approach. As noise is modeled at test time, unsupervised \ac{AVSE} can adapt to unseen noise situations and has a potentially better generalization performance than its supervised counterpart \cite{sadeghi2020audio}.

The AV-VAE models developed so far for unsupervised \ac{AVSE} do not account for the sequential nature of speech data, as they rely on a statistical independence assumption between consecutive speech time frames, and thus ignore their intrinsic correlations. Recently, some dynamical variants of VAEs, called DVAEs \cite{girin2021dynamical}, have been used for audio-only speech enhancement \cite{bie2022unsupervised}, which effectively model the temporal dynamics of speech data and improve the enhancement performance with respect to VAEs. Nevertheless, this comes at the cost of complicating the \ac{EM} step at test time, due to the complex temporal dependencies of latent variables in the models.  

In this paper, we extend the unsupervised \ac{AVSE} framework to audio-visual DVAE models, with a focus on respecting the computational efficiency of original non-sequential models for speech enhancement. To this end, we develop an audio-visual extension of the deep Kalman filter (DKF) model \cite{krishnan2017structured}, as the simplest DVAE variant in terms of temporal dependencies and architecture, which assumes a first-order Markov model on the latent variables. The proposed AV-DKF model efficiently incorporates visual data for clean speech generative modeling. Furthermore, we propose a dedicated \ac{EM}-based methodology for parameter learning and speech estimation at test time. Our experimental results demonstrate the superiority of the developed AV-DKF framework for speech enhancement compared with both its audio-only variant considered by Bie \textit{et al.} \cite{bie2022unsupervised}, and the standard AV-VAE model proposed in \cite{sadeghi2020audio}.

The rest of the paper is organized as follows. Section~\ref{sec:vae} reviews speech generative modeling and enhancement based on standard and dynamical VAEs. The proposed speech generative modeling and enhancement frameworks are detailed in Section~\ref{sec:prop}. Experimental results are then presented in Section~\ref{sec:exp}, followed by the conclusions in Section~\ref{sec:conc}.
\begin{figure*}[t]
 \centering
 \includegraphics[width=0.8\linewidth]{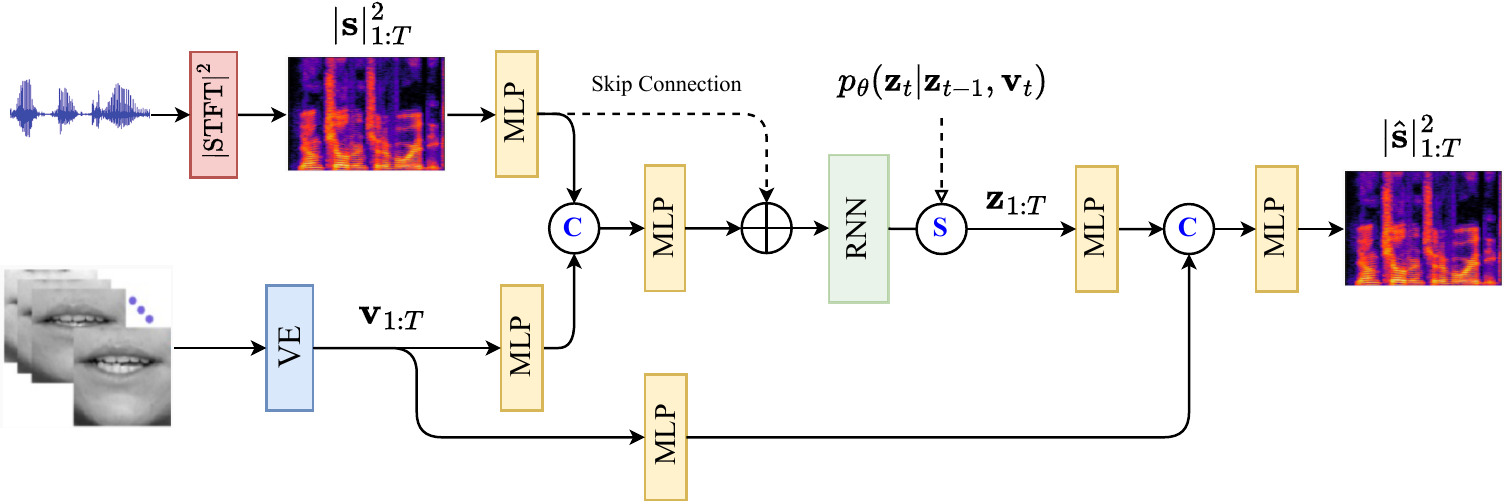}
 \caption{Schematic diagram of the proposed AV-DKF generative model (without explicit architecture of the prior network). \textbf{MLP}: multilayer perception, \textbf{RNN}: recurrent neural network, \textbf{VE}: video encoder, $\bigoplus$: addition, 
 {\textcolor{blue}{\textbf{C}}}: concatenation, \textcolor{blue}{\textbf{S}}: sampling in the latent space. }
 \label{fig:av_dkf}
\end{figure*}

\vspace{-2mm}
\section{Background}
\label{sec:vae}
Let us denote the \ac{STFT} representations of clean speech signals as $\vect{s}_{{1:T}}=\{\vect{s}_t\}_{t=1}^T$, where ${\vect{s}_t = [s_{ft}]_{f=1}^F \in \mathbb{C}^F}$. A latent variable $\vect{z}_t\in \mathbb{R}^L$ (${L\ll F}$) is associated to each \ac{STFT} time frame $\vect{s}_t$. The VAE framework then involves modeling the joint distribution of the observed and latent variables with some parametric Gaussian forms. The standard VAE model assumes the following factorization $p_\theta(\vect{s}_{{1:T}}, \vect{z}_{{1:T}}) = \prod_{t=1}^{T} p_\theta(\vect{s}_t, \vect{z}_t) = \prod_{t=1}^{T} p_\theta(\vect{s}_t| \vect{z}_t) p_\theta(\vect{z}_t)$. This modeling framework has been extended to the audio-visual case by conditioning the two distributions on visual features \cite{sadeghi2020audio}. No temporal modeling is considered here, which is not realistic for speech \ac{STFT} time frames. To resolve this issue, DVAEs consider the following factorization: $p_\theta(\vect{s}_{{1:T}}, \vect{z}_{{1:T}}) =  \prod_{t=1}^{T} p_\theta(\vect{s}_t| \vect{s}_{1:t-1}, \vect{z}_{1:t}) p_\theta(\vect{z}_t|\vect{s}_{1:t-1},\vect{z}_{1:t-1})$. The two distributions involved in this factorization are parameterized by some \ac{DNN} architectures, known as the \textit{decoder} and \textit{prior} networks, respectively.

Parameter inference, i.e., learning $\theta$, necessitates computation of the posterior distribution $p_\theta(\vect{z}_{{1:T}}| \vect{s}_{{1:T}})$, which is highly intractable due to the non-linear generative model. As a solution, a variational approximation is employed where a Gaussian form parameterized by a \ac{DNN}, called the \textit{encoder}, is introduced to approximate the intractable posterior \cite{KingW14}. For DVAEs, this writes $q_\psi(\vect{z}_{{1:T}}| \vect{s}_{{1:T}})=\prod_{t=1}^{T} q_\psi(\vect{z}_t|\vect{s}_{1:T},\vect{z}_{1:t-1})$. The inference phase involves joint learning of the model parameters by optimizing the so-called evidence lower bound (ELBO) of the intractable data log-likelihood $\log p_\theta(\vect{s}_{1:T})$ with stochastic gradient-based algorithms \cite{KingW14,bie2022unsupervised}. Depending on how the observed and latent variables are structured in the decoder and prior, several variants of the DVAE arise. In particular, DKF, as the simplest DVAE variant, assumes the following joint factorization $p_\theta(\vect{s}_{{1:T}}, \vect{z}_{{1:T}}) =  \prod_{t=1}^{T} p_\theta(\vect{s}_t| \vect{z}_{t}) p_\theta(\vect{z}_t|\vect{z}_{t-1})$, i.e., with a first-order Markov model on the latent variables.


The speech enhancement phase consists in combining the pre-trained speech generative model (the learned speech prior) with a parametric Gaussian model for noise, usually based on a \ac{NMF} variance model \cite{bando2018statistical}. The \ac{NMF} parameters are then learned from the observed noisy \ac{STFT} time frames, followed by speech signal estimation based on Wiener filtering. Here, one would also need to compute the posterior distribution of latent variables, which is intractable. A variational EM approach is proposed in \cite{bie2022unsupervised} that fine-tunes the pre-trained clean encoder on the noisy observations to approximate the intractable posterior. 

\vspace{-2mm}
\section{Proposed framework}
\label{sec:prop}
\subsection{Audio-visual DKF Generative model}
\label{sec:model}
We follow the DKF generative model \cite{girin2021dynamical}, and propose to extend it to the audio-visual case. Given some clean \ac{AV} training data $\vect{u}_{1:T} = \lk \vect{s}_{t},\vect{v}_{t}\rk_{t=1}^T$, with $\vect{v}_t$ being the visual feature vector at time frame $t$ extracted using a video encoder, the generative model is defined as follows:
\begin{equation}\label{eq:genmodel}
\begin{cases}
p_{\theta}(\vect{s}_t| \vect{z}_t, \vect{v}_t) = \mathcal{N}_c\Big(\bs{0}, \diag(\bs{\sigma}_{\theta_s}^2(\vect{z}_t, \vect{v}_t))\Big),\\
p_{\theta}(\vect{z}_t|\vect{z}_{t-1}, \vect{v}_t) = \mathcal{N}\Big(\bs{\mu}_{\theta_z}(\vect{z}_{t-1},\vect{v}_t), \diag(\bs{\sigma}_{\theta_z}^2(\vect{z}_{t-1},\vect{v}_t))\Big),\nonumber
\end{cases}
\end{equation}
where $\mathcal{N}_c(\bs{0}, \bs{\Sigma})$ denotes a circularly symmetric complex Gaussian distribution,  $\bs{\sigma}_{\theta_s}$, $\bs{\mu}_{\theta_z}$, $\bs{\sigma}_{\theta_z}^2$ are parametric non-linear functions realized by some \ac{DNN} architectures, and $\theta=\{ \theta_s, \theta_z\}$. The approximate posterior takes the following form:
\begin{align}
    q_\psi(\vect{z}_{1:T}|\vect{u}_{1:T}) = &\prod_{t=1}^{T} q_\psi(\vect{z}_t|\vect{r}_{t})\nonumber\\=&\prod_{t=1}^{T}\mathcal{N}\Big(\bs{\mu}_{\psi}(\vect{r}_{t}), \diag(\bs{\sigma}_{\psi}^2(\vect{r}_{t}))\Big)
\end{align}
where $\vect{r}_{t}=\lk \vect{z}_{t-1}, \vect{u}_{t:T} \rk$ collects all the conditioning variables, and $\bs{\mu}_{\psi}$, $\bs{\sigma}_{\psi}^2$ are \ac{DNN}-parameterized non-linear functions, i.e., the encoder. As with the previous works, the encoder takes the modulus square of \ac{STFT} data as input. Learning the set of parameters, i.e., $\Phi = \lk \theta, \psi \rk$, amounts to optimizing the ELBO:
\vspace{-0.2cm}
\begin{multline}\label{eq:elbo_av}
\mathcal{L}(\Phi;\vect{u}_{{1:T}}) = \sum_{t=1}^T \E_{q_{\psi}(\vect{z}_{t}| \vect{u}_{1:T}) }\lk\log p_{\theta}(\vect{s}_t| \vect{z}_t,\vect{v}_t)\rk - \\\sum_{t=1}^T \E_{q_{\psi}(\vect{z}_{t-1}| \vect{u}_{1:T}) }\lk\mathcal{D}_{\textsc{kl}}(q_\psi(\vect{z}_t|\vect{r}_t) \|p_\theta(\vect{z}_t|\vect{z}_{t-1}, \vect{v}_{t}))\rk,
\end{multline}
where $\mathcal{D}_{\textsc{kl}}(q \|p)$ denotes the Kullback–Leibler (KL) divergence between $q$ and $p$. The two expectations can be computed recursively, as detailed in \cite{girin2021dynamical}. A single-sample Monte-Carlo approximation of the two expectations is computed followed by the \emph{reparametrization trick} \cite{KingW14} before optimizing the parameters. The proposed AV-DKF architecture is shown in Fig.~\ref{fig:av_dkf}.
\subsection{Speech Enhancement}
\label{sec:se}
The observed noisy speech data are modeled as $ \vect{x}_t=\sqrt{g_t}\vect{s}_t+\vect{b}_t $, $t=1,\ldots,\tilde{T}$, where $ \vect{b}_t $ corresponds to noise. The parameters $\vect{g}_{1:\tilde{T}}=\lk{g_t}\rk_{t=1}^{\tilde{T}}$ are non-negative scalars to take into account the potentially different loudness between training and test speech data \cite{leglaive2018variance}. As the statistical model of $\vect{s}_t$, i.e., the prior distribution, the pre-trained AV-DKF generative model in \eqref{eq:genmodel} is used. Following the \ac{NMF} approach, two matrices 
$\Wb, \Hb$ of dimensions $F\times K$ and $K\times \tilde{T}$, respectively, with non-negative entries are considered for the variance of $ \bb_t $ as follows:
\begin{equation}
    \bb_t\sim \mathcal{N}_c(\boldsymbol{0}, \text{diag}(\Wb\bs{h}_t)),
    \label{eq:nmf}
\end{equation}
where $ \bs{h}_t $ is the $ t $-th column of $ \Hb $. As opposed to the previous works \cite{leglaive2018variance,bie2022unsupervised} that treat $\vect{g}_{1:\tilde{T}}$ as model parameters and estimate them using multiplicative update rules, here we propose a probabilistic modeling framework by assuming a gamma prior distribution for each $g_t$ as follows:
\begin{equation}\label{eq:gamma}
    p(g_t) = \frac{\beta^\alpha}{\Gamma(\alpha)} g_t^{\alpha-1} \exp(-\beta g_t),
\end{equation}
where $\Gamma(.)$ is the gamma function, and $\alpha, \beta>0$ (set to some predefined values), are the shape and scale parameters, respectively. As will be shown in Section~\ref{sec:exp}, this new approach results in significantly more stable and improved performance, especially for DKF-based models. Given the observed data $\vect{o}_{1:\tilde{T}} = \lk \vect{x}_t, \vect{v}_t \rk_{t=1}^{\tilde{T}}$, we follow an EM approach to estimate the set of model parameters $\phi=\lk \Wb, \Hb\rk$, which involves optimizing the expectation of the complete data log-likelihood with respect to the following posterior:
\vspace{-0.2cm}
\begin{multline}\label{eq:post_zw}
    p_\phi(\vect{z}_{1:{\tilde{T}}}, \vect{g}_{1:\tilde{T}}|\vect{o}_{1:\tilde{T}})\propto \\p_{\phi}(\vect{x}_{1:{\tilde{T}}}|\vect{z}_{1:{\tilde{T}}}, \vect{g}_{1:\tilde{T}},\vect{v}_{1:{\tilde{T}}}) p_\theta(\vect{z}_{1:{\tilde{T}}}|\vect{v}_{1:{\tilde{T}}}) p(\vect{g}_{1:\tilde{T}}),
\end{multline}
where the likelihood writes:
\vspace{-0.5cm}
\begin{multline}\label{eq:likelihood}
    p_{\phi}(\vect{x}_{1:{\tilde{T}}}|\vect{z}_{1:{\tilde{T}}}, \vect{g}_{1:\tilde{T}},\vect{v}_{1:{\tilde{T}}}) = \prod_{t=1}^{\tilde{T}}p_{\phi}(\vect{x}_t|\vect{z}_t, {g}_t, \vect{v}_t) \\= \prod_{t=1}^{\tilde{T}}\mathcal{N}_c\Big(\bs{0}, \diag(g_t\bs{\sigma}_{\theta_s}^2(\vect{z}_t, \vect{v}_t) + \Wb\bs{h}_t)\Big).
\end{multline}
Unfortunately, there is no closed form expression for \eqref{eq:post_zw}. However, as an efficient approximate approach inspired by \cite{kameoka2019supervised,leglaive2020recurrent}, we try to find the mode of the posterior distribution of $\vect{z}_{1:{\tilde{T}}}, \vect{g}_{1:\tilde{T}}$:
\vspace{-0.25cm}
\begin{multline}\label{eq:zw}
    \vect{z}_{1:\tilde{T}}^*, \vect{g}_{1:\tilde{T}}^* = \argmax_{\vect{z}_{1:T}, \vect{g}_{1:\tilde{T}}}~\sum_{t=1}^{\tilde{T}}\log p_{\phi}(\vect{x}_t|\vect{z}_t, {g}_t, \vect{v}_t)+\\\log p_{\theta}(\vect{z}_t|\vect{z}_{t-1}, \vect{v}_t) + \log p(g_t),
\end{multline}
which, after substituting from \eqref{eq:genmodel}, \eqref{eq:gamma}, and \eqref{eq:likelihood}, can be optimized by a few iterations of a gradient-based solver, e.g., Adam \cite{KingmaB15}. In the maximization (M) step, the \ac{NMF} parameters are updated according to the following approximate problem:
\begin{equation}\label{eq:wh}
    \Wb, \Hb \leftarrow \argmax_{\Wb, \Hb}~ \sum_{t=1}^{\tilde{T}} \log p_{\phi}(\vect{x}_t|\vect{z}_t^*, {g}_t^*, \vect{v}_t),
\end{equation}
which can be solved with multiplicative update rules as similarly done in \cite{leglaive2018variance}. The overall inference algorithm iterates between \eqref{eq:zw} and \eqref{eq:wh}. Once $\phi^*=\lk\Wb^*,\Hb^*\rk$ is learned, the speech signal is estimated as the posterior mean (element-wise division):
\begin{align}\label{eq:wf}
    \hat{\vect{s}}_{1:\tilde{T}} &= \E_{p_{\phi^*}(\vect{s}_{1:\tilde{T}}|\vect{o}_{1:\tilde{T}})}\lk \vect{s}_{1:\tilde{T}}\rk\nonumber\\
    &=\E_{p_{\phi^*}(\vect{z}_{1:\tilde{T}}^*, \vect{g}_{1:\tilde{T}}^*|\vect{o}_{1:\tilde{T}})}\lk \E_{p_{\phi^*}(\vect{s}_{1:\tilde{T}}|\vect{z}_{1:\tilde{T}}^*, \vect{g}_{1:\tilde{T}}^*, \vect{o}_{1:\tilde{T}})}\lk \vect{s}_{1:\tilde{T}}\rk\rk\nonumber\\
    &\approx\lk\frac{g_t^*\bs{\sigma}_{\theta}^2(\vect{z}_t^*, \vect{v}_t)}{g_t^*\bs{\sigma}_{\theta}^2(\vect{z}_t^*, \vect{v}_t) + \Wb^*\bs{h}_t^*}\odot \vect{x}_t \rk_{t=1}^{\tilde{T}}.
\end{align}
\begin{table*}[t!]
\centering
\caption{Average values of the SI-SDR, PESQ, and STOI metrics for the input (unprocessed) and output (enhanced) test speech signals. For each method, \textbf{top row}: $g_t$ updated by multiplicative rules \cite{leglaive2018variance,bie2022unsupervised}, \textbf{bottom row}: $g_t$ updated according to \eqref{eq:zw}.}\vspace{-2mm}
\resizebox{\textwidth}{!}{
\begin{tabular}{|l|c|c|c|c|c||c|c|c|c|c||c|c|c|c|c|}
\hline
 Metric & \multicolumn{5}{c||}{SI-SDR (dB)} & \multicolumn{5}{c||}{PESQ} & \multicolumn{5}{c|}{STOI} \\
\hline
{SNR (dB)} & {-5} & {0} & {5} & {10} & {15} & {-5} & {0} & {5} & {10} & {15} & {-5} & {0} & {5} & {10} & {15} \\ \hline\hline
Input &-12.80 & -7.72 & -2.91  & 2.04 & 7.25 &  1.51 & 1.76 & 2.05 & 2.37 & 2.85 & 0.20       & 0.30       & 0.43       & 0.56 & 0.69                    \\ \hline\hline
\multirow{ 2}{*}{A-VAE} & \textbf{-7.37} & \textbf{-1.92} & \textbf{3.78} & \textbf{8.65} & \textbf{13.07} & 1.63 & 1.91 & 2.20 & 2.50 & 2.85 & 0.21         & 0.32        & 0.45      & 0.59       & 0.72 \\ \cline{2-16}
& -8.46 & -2.60 & 3.02 & 8.11 & 13.01  & \textbf{1.67} & \textbf{1.95} & \textbf{2.25} & \textbf{2.58} & \textbf{2.90} & \textbf{0.22} & {{0.32}} & {\textbf{0.47}} & {\textbf{0.60}} & {\textbf{0.73}}\\ \hline\hline
\multirow{ 2}{*}{AV-VAE} & -6.86 & \textbf{-0.83} & \textbf{4.70} & \textbf{9.38} & \textbf{13.90} & 1.74 & 2.00 & 2.31 & 2.61 & 2.90  & 0.20                    & 0.31                  & 0.45                     & 0.59                     & 0.72 \\ \cline{2-16}
 & \textbf{-6.65} & {-0.86} & {4.47} & {9.26} & {13.77} & \textbf{1.75} & \textbf{2.03} & \textbf{2.34} & \textbf{2.65} & \textbf{2.93} & {\textbf{0.22}} & {\textbf{0.33}}  & {\textbf{0.47}} & {\textbf{0.61}}  & {\textbf{0.73}} \\ \hline\hline
\multirow{ 2}{*}{A-DKF} &  \textbf{-6.50} & -1.41 & 1.99 & 4.36 & 5.55 & 1.48 & 1.67 & 1.87 & 2.02 & 2.13 & 0.22                    & 0.33                  & 0.45                     & 0.55                     & 0.64 \\ \cline{2-16}
 & {-7.02} & \textbf{-0.92} & \textbf{4.76} & \textbf{10.39} & \textbf{14.96} & \textbf{1.78} & \textbf{2.08} & \textbf{2.41} & \textbf{2.75} & \textbf{3.03} & {{0.22}} & {\textbf{0.35}}  & {\textbf{0.50}} & {\textbf{0.65}}  & {\textbf{0.77}} \\ \hline\hline
\multirow{ 2}{*}{AV-DKF} & -5.04 & -0.21 & 2.93 & 4.92 & 5.48  & 1.39 & 1.61 & 1.82 & 1.97 & 2.07 & 0.22                    & 0.33                  & 0.44                     & 0.55                     & 0.63 \\ \cline{2-16}
 & \textbf{-3.78} & \textbf{1.78} & \textbf{7.19} & \textbf{11.66} & \textbf{15.81} & \textbf{1.94} & \textbf{2.24} & \textbf{2.54} & \textbf{2.80} & \textbf{3.05} & {\textbf{0.25}} & {\textbf{0.38}}  & {\textbf{0.52}} & {\textbf{0.66}}  & {\textbf{0.77}} \\ \hline
\end{tabular}}
\label{tab:se_results}\vspace{-3mm}
\end{table*}
\vspace{-.5cm}
\section{Experiments}
\label{sec:exp}
In this section, we provide a performance evaluation of our proposed AV-DKF speech enhancement framework against some baseline methods, including A-VAE \cite{leglaive2018variance}, AV-VAE \cite{sadeghi2020audio}, and A-DKF \cite{bie2022unsupervised}. To measure the quality of the enhanced speech signals, we use standard metrics, including the scale-invariant signal-to-distortion ratio (SI-SDR) in dB \cite{le2019sdr}, the short-term objective intelligibility (STOI) measure~\cite{Taal2011stoi}, ranging in $[0,1]$, and the perceptual evaluation of speech quality (PESQ) score~\cite{Rix2001pesq}, ranging in $[-0.5,4.5]$. For all the metrics, the higher, the better.

\noindent\textbf{Datasets}. For training all the VAE variants, we used the TCD-TIMIT corpus \cite{harte2015tcd}. This dataset contains \ac{AV} speech data from 56 English speakers (39 for training, 8 for validation, and 9 for testing) with an Irish accent, uttering 98 different sentences, each $\sim$ 5-second long, and sampled at 16 kHz. This amounts to $\sim$ 8 hours of data. There is a corresponding video file for each utterance that captures a frontal view of the speaker at a rate of 30 frames per second. For all the videos, the lip \ac{ROI} are already extracted as $67\times 67$ images (a sample is shown in Fig.~\ref{fig:av_dkf}). The STFT of the speech data is computed with a 1024 samples-long (64 ms) sine window, 75$\%$ overlap, without zero-padding, yielding STFT frames of length $F=513$. The \ac{ROI} images for each video were upsampled across time to make visual and audio frame rates equal. Moreover, we augmented each video into 75 different samples by applying natural image transformations such as random translation up to 10\%, random scaling and crop up to 10\%, and random brightness and contrast jitter up to 40\%.

To test the speech enhancement performance, we used the noisy speech material of the NTCD-TIMIT corpus \cite{abdelaziz2017ntcd} which has been created by adding six noise types, including \textit{\ac{LR}}, \textit{White}, \textit{Cafe}, \textit{Car}, \textit{Babble}, and \textit{Street}, with different signal-to-noise (SNR) ratios to
the test speech data of the TCD-TIMIT corpus. We randomly selected 5 utterances per noise level and noise type from each test speaker, which resulted in 1350 test utterances.

\noindent\textbf{Models architectures}. The A-VAE and AV-VAE models share the same architectures as the baseline VAE model experimented in \cite{leglaive2020recurrent,bie2022unsupervised}, where the encoder and decoder comprise a single fully connected (FC) hidden layer with 128 nodes and tanh activation functions. For the DKF models, we followed a similar architecture as the one proposed in \cite{bie2022unsupervised}, which consists of a backward long short-term memory (LSTM) network, a combiner function in the encoder, and a gated transition function in the prior network \cite{krishnan2017structured}. The decoder comprises a multilayer perception (MLP) with four hidden layers of dimensions 32, 64, 128, and 256, with the tanh activation functions. We took the pre-trained A-VAE and A-DKF models of \cite{bie2022unsupervised}, trained on the (audio-only) Wall Street Journal (WSJ0) corpus \cite{garofolo1993wsj0}, and fine-tuned them on our training data.\footnote{\url{https://github.com/XiaoyuBIE1994/DVAE_SE}} Similarly, we fine-tuned AV-VAE and AV-DKF from their audio-only pre-trained counterparts \cite{bie2022unsupervised}.

For the AV models, the raw visual data are processed by a feature extraction network, called VE (video encoder) in Fig.~\ref{fig:av_dkf}, before feeding to the model. For that, we incorporated a pre-trained model that is part of the lipreading network proposed in \cite{martinez2020lipreading}. We took the initial blocks of the visual network, which include a 3D convolutional module and a ResNet architecture module. We used the computed features as the visual input of our model, without fine-tuning the VE network during the training process. The introduced ``skip connection'' in Fig.~\ref{fig:av_dkf} helps stabilize the contribution of visual information, since parts of the model are fine-tuned from A-DKF.

\noindent\textbf{Parameters settings}. For all the models, the latent dimension is set to $L=16$. Moreover, the NMF parameters are initialized with non-negative random entries (the same values for all the methods), with $K=8$. For the DKF models, we used a sequence length of $T=50$, as in \cite{bie2022unsupervised}. Also, we set the gamma parameters in \eqref{eq:gamma} to $\alpha=\beta=1$, and initialized $\vect{g}_{1:\tilde{T}}$ with an all-one vector. Furthermore, $\vect{z}_{1:\tilde{T}}$ is initialized by feeding $\vect{x}_{1:\tilde{T}}$ (and $\vect{v}_{1:\tilde{T}}$, for \ac{AV} models) to the encoder and taking the mean of $q_\psi$. All the models are trained with the Adam optimizer, with a learning rate of 0.0001 and a batch size of 128. We used early stopping on the validation set with a patience of 50 epochs, i.e., the training stops if the validation loss does not improve after 50 consecutive epochs. For all the models, the number of EM iterations for speech enhancement was set to 100, where at each iteration, the E-step \eqref{eq:zw} was performed using the Adam optimizer, for 20 iterations and with a learning rate of $0.001$.

\begin{figure}[t]
 \centering
\includegraphics[width=1\linewidth]{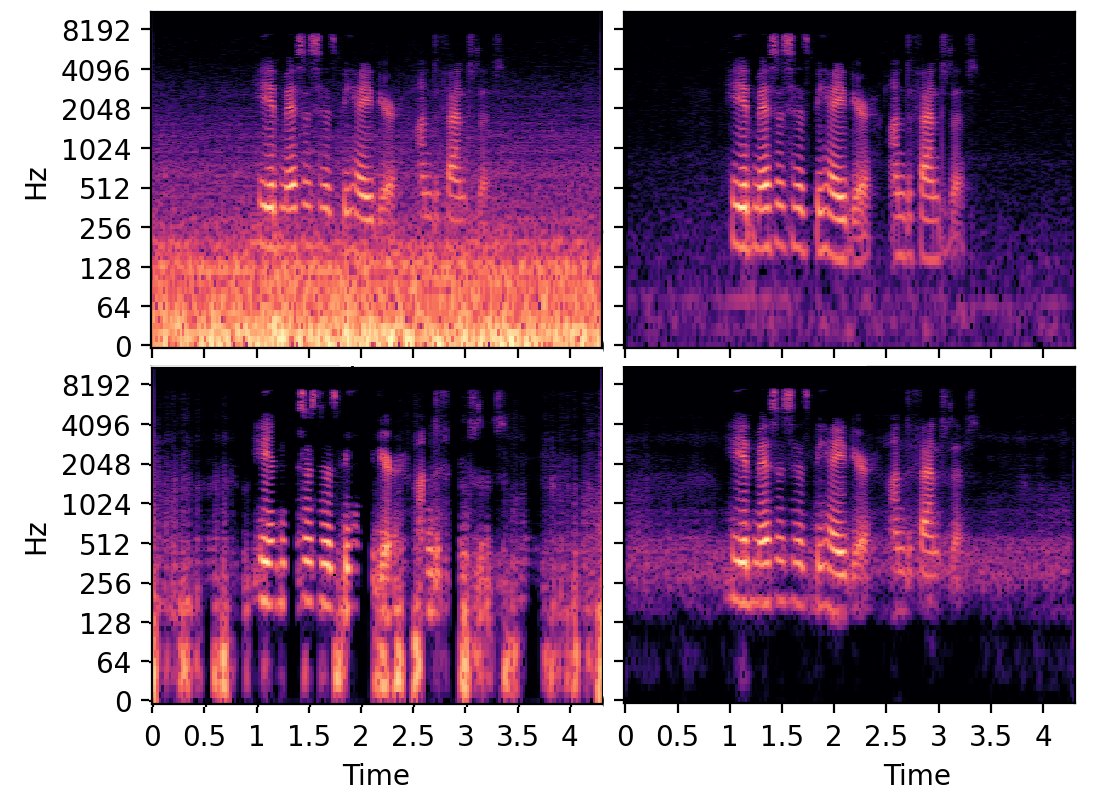}
\vspace{-0.6cm}
\caption{Effect of $g_t$ update on AV-DKF speech enhancement. From top to bottom, left to right: noisy, clean, output of multiplicative update rule \cite{leglaive2018variance,bie2022unsupervised}, output of optimization-based update rule \eqref{eq:zw}.}
\label{fig:example}
\end{figure}

\noindent\textbf{Results}. The speech enhancement results are reported in Table~\ref{tab:se_results}, where for each competing method, we report two sets of results corresponding to the two different approaches for updating the scaling parameter $g_t$ in the observation model. In each cell, the top and bottom rows correspond, respectively, to the multiplicative update rule proposed in \cite{leglaive2018variance,bie2022unsupervised} and the probabilistic framework proposed in this work, i.e., \eqref{eq:gamma} and \eqref{eq:zw}. By inspecting Table~\ref{tab:se_results} we can draw several conclusions. First, as can be clearly seen, the proposed update framework for $g_t$ yields more stable and improved performance than the multiplicative updates, especially for the DKF models. Specifically, for the non-sequential VAEs, i.e., A-VAE and AV-VAE, the PESQ and STOI values consistently improved but for SI-SDR there is a degradation, which is less significant for AV-VAE. The effect of the proposed update rule is more noticeable for the DKF models, without which, the models do not work well, especially for higher noise levels. In fact, we noticed in our experiments that without constraining $g_t$ with a proper regularization, e.g., the gamma prior, the enhancement method for the DKF models would lead to improper noise removal. Fig~\ref{fig:example} shows an illustrative example with AV-DKF, where we can see that some speech time frames are wrongly estimated either as zero or noisy, which could be due to too small and too large values for $g_t$, respectively (cf. Equation \eqref{eq:wf}). 

We also see a clear and consistent performance improvement for the DKF models compared with their non-sequential versions. Specifically, AV-DKF exhibits an average performance gain of about \textbf{2.5 dB} in SI-SDR, \textbf{0.18} in PESQ, and \textbf{0.05} (5\%) in STOI over AV-VAE. This signifies the importance of temporal modeling. Comparing the results of AV-DKF and its audio-only version, A-DKF, demonstrates average gains of about \textbf{2 dB} in SI-SDR, \textbf{0.10} in PESQ, and \textbf{0.02} (2\%) in STOI. This proves the usefulness of visual information for speech enhancement. Moreover, the amount of improvement is higher for larger amounts of noise, i.e., situations wherein the role of visual modality is more highlighted, meaning that AV-DKF is able to efficiently incorporate the useful information of visual data for speech enhancement. Supplementary material, including audio-visual examples, will be available online.\footnote{\url{https://team.inria.fr/multispeech/demos/av-dkf/}}
\vspace{-5mm}
\section{Conclusion}\label{sec:conc}
\vspace{-2mm}
We presented the audio-visual deep Kalman filter (AV-DKF) model to learn the prior distribution of clean speech data for speech enhancement. In contrast to the non-sequential models used in the prior work, the AV-DKF model incorporates visual data more efficiently. Besides, we developed an inference algorithm for speech enhancement based on the learned speech prior. Our experiments confirmed the superiority of the AV-DKF model compared with its audio-only version and non-sequential models. Future work includes extending the proposed framework to other dynamical models \cite{girin2021dynamical}. 

\bibliographystyle{IEEEbib}
\bibliography{mybib}

\end{document}